\documentclass[sigconf]{acmart}

\usepackage{booktabs} 
\usepackage{blindtext}
\usepackage{subcaption}

\acmDOI{10.475/123_4}

\acmISBN{123-4567-24-567/08/06}

\acmConference[]
\acmYear{2018}

\begin{document}
\title{Accelerating Prototype-Based Drug Discovery using Conditional Diversity Networks}

\author{Shahar Harel}
\affiliation{%
  \institution{Technion - Israel Institute of Technology}
  \city{Haifa} 
  \state{Israel} 
}
\email{sshahar@cs.technion.ac.il}

\author{Kira Radinsky}
\affiliation{%
  \institution{Technion - Israel Institute of Technology}
  \city{Haifa} 
  \state{Israel} 
}
\email{kirar@cs.technion.ac.il}

\renewcommand{\shortauthors}{S. Harel et al.}

\begin{abstract}
Designing a new drug is a lengthy and expensive process. 
As the space of potential molecules is very large ($10^{23}-10^{60})$, a common technique during drug discovery is to start from a molecule which already has some of the desired properties. 
An interdisciplinary team of scientists generates hypothesis about the required changes to the prototype.
In this work, we develop an algorithmic unsupervised-approach that automatically generates potential drug molecules given a prototype drug.
We show that the molecules generated by the system are valid molecules and significantly different from the prototype drug.
Out of the compounds generated by the system, we identified 35 FDA-approved drugs.
As an example, our system generated Isoniazid -- one of the main drugs for Tuberculosis. 
The system is currently being deployed for use in collaboration with pharmaceutical companies to further analyze the additional generated molecules.
\end{abstract}

%
%

\keywords{Prototype-Based Drug Discovery, Drug Design, Deep Learning for Medicine}

\maketitle

\newcommand*{\NAME}{{\em CDN }}

\section{Introduction}

Producing a new drug is an expensive and lengthy process that might take over 500 million dollars and over 10--15 years.
The first stage is drug discovery, in which potential drugs are identified before selecting a candidate drug to progress to clinical trials. 
Although historically, some drugs have been discovered by accident (e.g., Minoxidil and Penicillin), today more systematic approaches are common.
The most common method involves screening large libraries of chemicals in high-throughput screening assays (HTS) to identify an effect on potential targets (usually proteins). The goal of such a process is to identify compounds that might modify the target activity, which might often result in a therapeutic effect.


While HTS is a commonly used method for novel drug discovery, it is common to start from a molecule which already has some of the desired properties. Such a molecule, usually called a ``prototype'', might be extracted from a natural product or a drug on the market which could be improved upon. 
Intuitively, producing a chemically and structurally related substance to an existing active pharmaceutical compound usually improves on the efficacy of the prototype drug -- 
reduces adverse effects, works on patients that are resistant to the prototype, and might be less expensive \cite{garattini1997me}.

During this process of prototype-based drug discovery, an interdisciplinary team of scientists generates hypothesis about the required changes to the prototype.
One might consider this process as a pattern recognition process -- chemists, through their work, gain experience in identifying correlations between chemical structure retrosynthetic routes and pharmacological properties \cite{naturepaper}.
They rely on their expertise and medicinal chemistry
intuition to create chemical hypotheses, which have been shown to be biased \cite{schnecke2006computational}.
However, the chemical space is virtually infinite --  the amount of synthetically valid
chemicals which are potentially drug-like molecules is estimated to be between $10^{23}-10^{60}$ \cite{sizeDrugLike}.
In this work, we develop an algorithmic unsupervised approach to automatically generate potential drug molecules given a prototype drug. 

It is common to encode molecular structures into SMILES notations (simplified molecular-input line-entry system) that preserves the chemical structural information.
For example, Methyl isocyanate can be encoded using the following string: CN=C=O.
We learn embeddings of drug-like molecules in molecule space represented by SMILES. 
To identify drug-like molecules, which are used to train our algorithm, we use the Lipinski criteria -- a common chemical drug-design qualitative measure that estimates the structure bioavailability, solubility in water, potency, etc. \cite{lipinski} 

\begin{figure}
\includegraphics[width=0.5\textwidth]{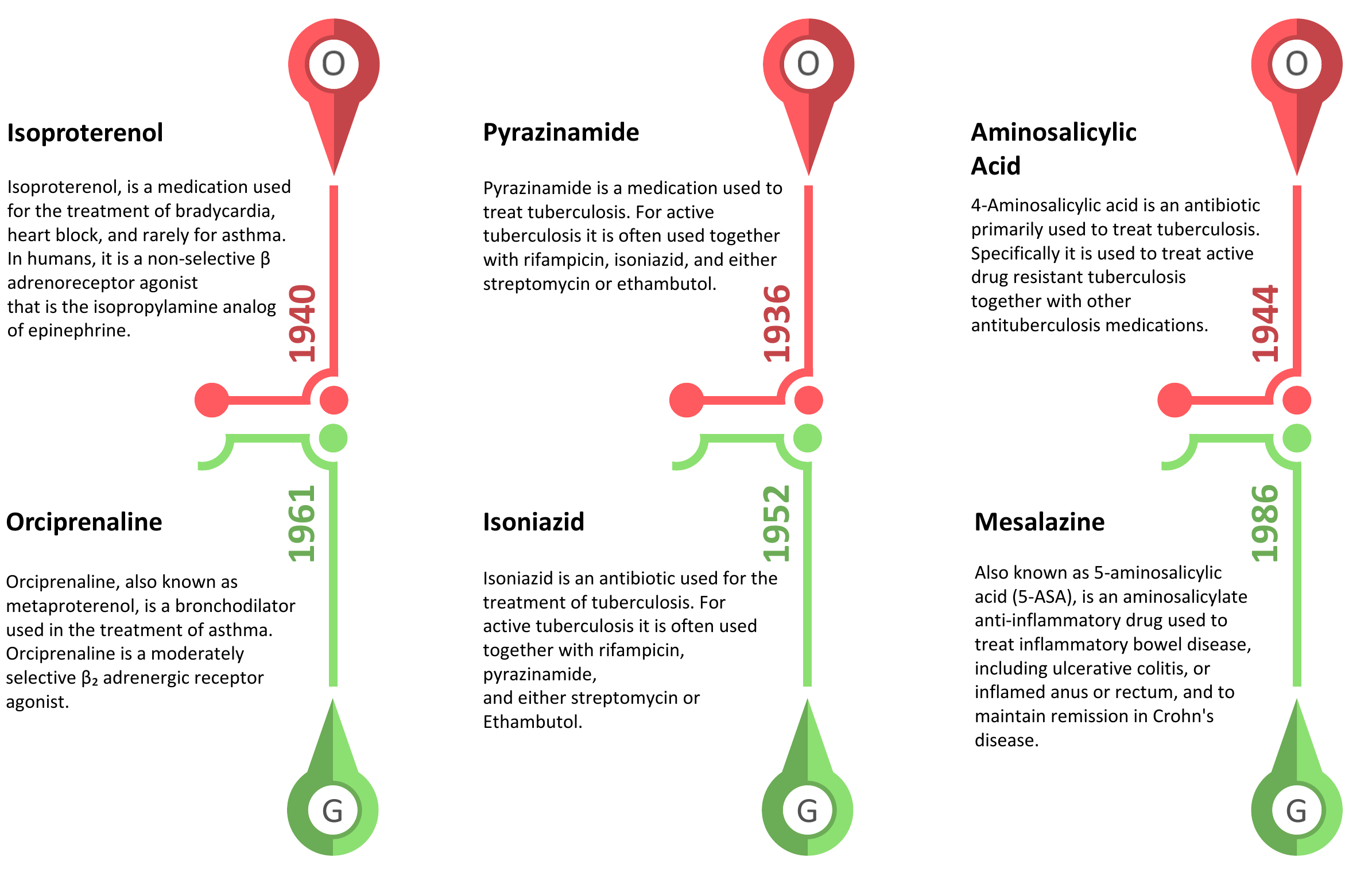}
\caption{Drug development timeline, with example of drugs generated by \NAME (bottom), using FDA approved drugs as prototypes (top).}
\vspace{-0.3cm}
\label{fig:timeline}
\end{figure}

Variational Auto Encoders (VAE) \cite{kingmaVAE} are encoder-decoder architecture that attempts to learn the data distribution in a way that can later be sampled from to generate new examples. State-of-the-art results have been shown for generating images that resemble natural images, yet not identical to the train data \cite{kingmaVAE, larsenVAE}. Training a vanilla VAE on drug-like molecules provides an ability to sample new molecules which intuitively should be drug-like \cite{automaticChemicalDesign}. 
In this work, we extend VAE to allow a conditional sampling -- sampling an example from the data distribution (drug-like molecules) which is closer to a given input. This allows sampling molecules closer to a prototype drug, and thus increase probability of generating a valid drug with similar characteristics. Additionally, we add a diversity component that allows the sampling to be different from the prototype drug as well.
We present a deep-learning approach which we call Conditional Diversity Networks (CDN), which allows the diverse conditioned sampling.
The results show that the molecules \NAME generates are similar to the prototype drugs yet significantly diverse. We show empirical results that the system generates high percentage of valid molecules.
Additionally, 
we perform retrospective experiments and use drugs developed in the 1930's and 1940's as prototypes. The system was then able to generate new drugs, some of which discovered dozens of years after the prototype discovery (Figure \ref{fig:timeline}). 

One such example is the system discovery of the main drug for Tuberculosis -- Isoniazid. Discovered in 1952, it is on the World Health Organization's List of ``Essential Medicines, the most effective and safe medicines needed in a health system'' \cite{WHOiso}.
In the retrospective experiment, we used as prototypes only drugs discovered until 1940.
For the drug Pyrazinamide, first discovered in 1936, the system generated the SMILES notation of what today is known as Isoniazid.
Pyrazinamide, although discovered in 1936, was not used until 1972 for treating Tuberculosis. 
Tuberculosis can become resistant to treatment if Pyrazinamide is used alone and therefore is always used in combination with Isoniazid and others. The combination reduces treatment time from 9 months to  
less than 3 months.
This example shows promise on how substances that could not be used at the time of discovery can serve as a prototype for discovering new drugs.
In collaboration with pharmaceutical companies additional generated molecules are being tested today.
We believe our system lays the foundations to build algorithmically-directed HTS based on prototype drugs.

\section{Related Work} \label{relatedWork}
Over the past decade, deep neural networks (DNN) has been a game changer in various areas of machine learning research, such as computer vision \cite{Krizhevsky,vision2}, natural language processing \cite{mikolov1} and speech recognition \cite{hintonSpeech}. Deep neural network most prominent success stories are observed in domains with access to large, raw (unprocessed) datasets. In such scenarios deep learning was able to achieve above human level performance.
Compared with those domains, DNN in chemistry relies heavily on engineered features representing a molecule \cite{feature3massively, feature4Convolutional, feature2deeptox}. Such approaches are semi optimal as they restrict the space of potential representations through the assumptions made by limiting to the chosen features \cite{chemnet}.

More recent methods overcome this issue by leveraging advanced deep neural network models to learn chemical continuous representations (i.e., embeddings) based on a large datasets of molecular raw data. Molecular raw data can be represented in few ways, and processed with different deep architectures. Among those we can find 2D/3D images served as input to a convolutional neural network (CNN) \cite{Atomnet, chemnet}, molecular graph representation paired with neural graph embedding methods \cite{convolutionalGraph, moleculenet}, and SMILES strings -- modeled as a language model with recurrent neural network (RNN) \cite{FoundinTranslation, automaticChemicalDesign, smilesEnumeration}.
\citeauthor{organic_Weisfeiler, feature4Convolutional, FoundinTranslation} leverage the embeddings for numerous supervised prediction task, e.g. predicting outcomes of complex organic chemistry reactions. 

Recently, deep generative models have opened up new avenues for leveraging molecular embeddings for unsupervised tasks such as molecule generation, and drug design.
Most methods aim at generating valid molecules.
For example \citeauthor{segler2017generating} train RNN as a language model to predict the next character in a SMILES string. After training, the model can be used to generate new sequences corresponding to new molecules. \citeauthor{automaticChemicalDesign} leverage on the VAE \cite{kingmaVAE} generative model, to learn dense molecular embedding space. At test time, the model is able to generate new molecules from samples of the prior distribution enforced on the latent representation during training.
In this general form of generation, we can only hope to achieve the task of generating molecule libraries with no specific chemical/biological characteristics, but the characteristics of the training data.
Others extend this approach by tuning the model on a dataset of molecules with specific characteristics \cite{segler2017generating}, or by applying post processing methods, such as Bayesian Optimization \cite{Bayesianmolecular,automaticChemicalDesign} and Reinforcement Learning \cite{reinforcementMolecular}.

In this work, we target the problem of generating drug-like molecules and show that training vanilla generative models on this family shows limited results (Section \ref{s:Experiments}), both for generating diverse novel molecules and for generating drugs.
Following the common chemical approach, we focus the generative approach on a given prototype. This helps ``guide'' the search process  around the prototype in the chemical space.
Given prototypes can be drug-like molecules or known drugs.
We introduce parametrized diversity
and design an end-to-end neural network solution to train the model to represent the chemical space, and to allow for further diversity driven prototype based exploration and novel molecule generation.

\section{Methods}
We define the problem of prototype-driven hypothesis generation as a conditional data generation process. The model operates on a given molecule prototype and generates various molecules as candidates. The generated molecules should be novel and share desired properties with the prototype. 
The main contribution of our work is enabling prototype-based generation with a diversification factor.
We start by reviewing how molecules are represented as text (Section \ref{moleculeRepresentation}) and then present a generative model (Section \ref{dataGeneration}).
Our generative model builds upon recent methods for deep representation learning. We train a stochastic neural network to learn internal molecule representation (embedding). After obtaining the molecule embedding we further utilize the stochastic component of the neural architectures to introduce parametrized diversity layer into the generation process. 
The architecture of our proposed solution is presented in Section \ref{Architecture}.

\subsection{Molecule Representation} \label{moleculeRepresentation}
The choice of representation of molecules is at the heart of any computer-based chemical analysis. 
For molecule generation, it is of crucial importance, as the task is to both analyze and generate objects of the same representation.
\citeauthor{smilesNLP} showed that organic molecules contain fragments whose rank distribution is essentially identical to that of sentence fragments. The consequence of this discovery is that the vocabulary of organic chemistry and human language follow very similar laws. Intuitively, there is an analogy between a chemist understanding of a compound and a language speaker understanding of a word. This introduces a potential to leverage recent advances in linguistics-based analysis, and deep sequence models in particular.

A SMILES string is a commonly-used text encoding for organic molecules. SMILES represents a molecule as a sequence of characters corresponding to atoms as well as special characters denoting opening and closure of rings and branches.
For example c and C represent aromatic and aliphatic carbon atoms, O represents oxygen, -, = and \# represent single, double and triple bonds \cite{smiles}. Then a molecules, such as Benzene, is represented in SMILES notation as c1ccccc1.
It has already been shown that SMILES representation of molecules has been effective in chemoinformatics \cite{FoundinTranslation,automaticChemicalDesign, smilesEnumeration, segler2017generating}. This has strengthened our belief that recent advances in the field of deep computational linguistics and generative models might have an immense impact on prototype based drug development.


\begin{figure*}
\includegraphics[width=1.0\textwidth,height=0.35\textwidth]{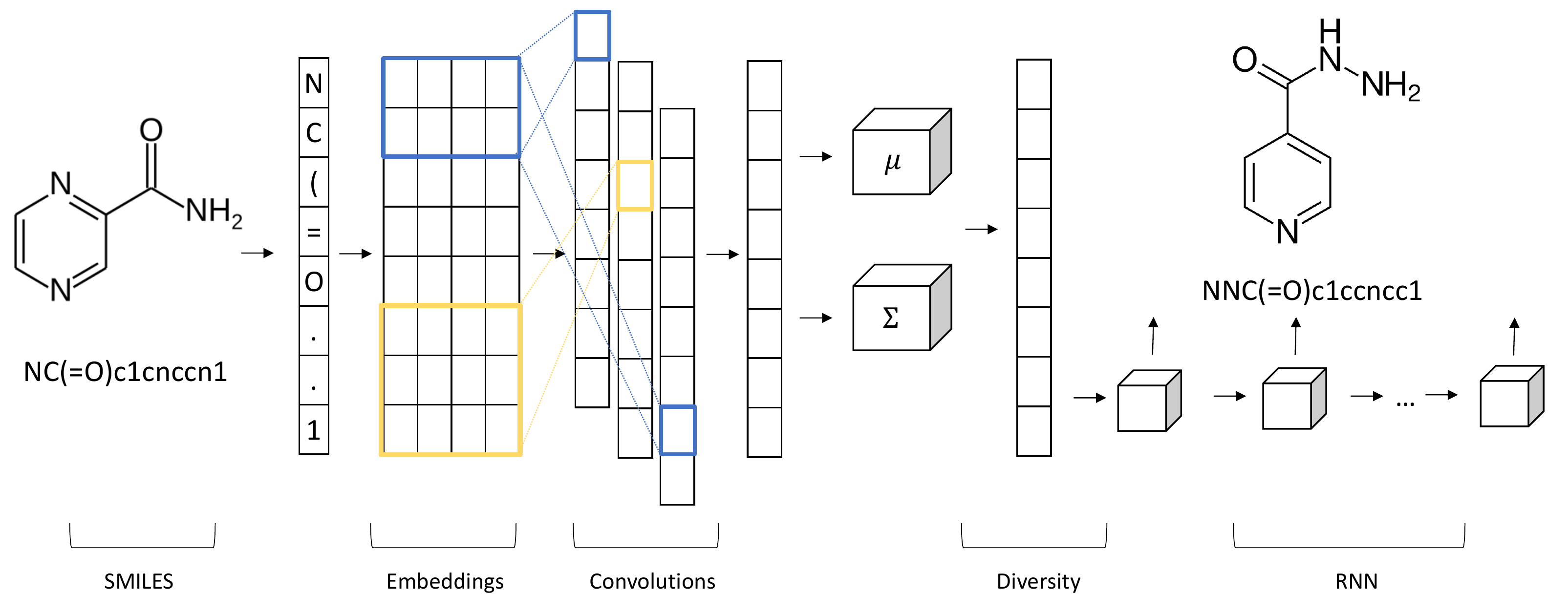}
\caption{\NAME end-to-end neural net architecture}
\label{fig:architecture}
\end{figure*}

\subsection{Molecule Driven Hypothesis Generation} \label{dataGeneration}
Generative models have been applied for many tasks, e.g., image generation.
The models synthesized new images which resembled the database the models were trained on \cite{kingmaVAE,larsenVAE}.
One of the most popular generative frameworks are Variational Autoencoders (VAE) \cite{kingmaVAE}.
VAE  are encoder-decoder models that use a variational approach for latent representation learning.
Intuitively, the encoder is constrained to generate latent representations that follow a prior.
During generation, latent vectors are sampled from the priors and passed to the decoder that generates the new representation.
We leverage VAE for the task of molecule generation. The stochasticity allows integrating chemical diversity into the generation process.
However, application of generative models for molecule generation have shown limited results \cite{automaticChemicalDesign}. Unlike image generation, where each image is valid, when we aim at molecule generation, not each representation is a valid molecule representation. 
Intuitively, when we sample from the prior for image generation -- the space of images is much more dense than that of valid molecules. Therefore, many image samples are valid compared to randomly generated molecules representations. 
We hypothesize that a constrained generation next to a known prototype, rather than a non-constrained sampling, will yield better molecule generation.
We  extend VAE generation process to condition on a prototype, i.e., generate molecules closer to a given drug. 
Intuitively, directing the sampling process closer to existing prototype drugs might yield valid molecules that carry similar characteristics to the the prototype yet provide diversity.
Our results provide evidence that a conditioned sample along side a diversity component yields more valid and novel results. If conditioned on known drugs, the system is able to generate drugs discovered years after the prototype (Section \ref{dataGeneration}).

More formally, we assume a molecule $M$ has a latent representation $z$ that captures the main factors of variation in the chemical space. We model the covariates $z_{i} \mid M$ as Gaussians ($z_{i} \sim \mathcal{N}(\mu_{i},\sigma_{i})$).
With the latent representation $z$ at hand, we want to generate a candidate molecule in a SMILES discrete form, therefore we define the generative model $ \hat{y} \mid z \sim Multi(\theta) $,
where $y$ is the generated candidate, Formally:
\vspace{-0.2cm}
\begin{equation}
q(z \mid M) = \prod_{i=1}^{D_z}{\mathcal{N}(\mu_{i}= \hat{\mu_{i}},\sigma{i}=\hat{\sigma{i}}}) 
\end{equation}
\vspace{-0.2cm}
\begin{equation}
p(\hat{Y} \mid z) = Multi(\theta = \hat{\theta}) 
\end{equation}

Where q is approximated via encoder neural network function, applied on molecule $M$ as input, and outputs the latent feature parameterization ($\hat{\mu}, \hat{\sigma}$) of the molecule. We than sample instance from this parametrization to obtain the final encoded output $z$;
p is represented via decoder neural network, applied on the molecule sampled feature instance $z$ as input, and generates the output molecule as described below.

Generating a molecule as a SMILES string reflects multinomial distribution over the atoms space. Each atom is represented via a character.
We form the character generation process as an iterative process, each character $y_i$ is generated based the the hidden encoded representation $z$ and the formerly generated $y_i$'s. In total, the output of this step is a string $\hat{y_i}=\lbrace \hat{y_1},..,\hat{y_N} \rbrace $, where $N$ is a pre-defined maximal generation length. Formally, for a single character $y_i$

\begin{equation}
P(\hat{y_i} \mid \hat{y_1},\ldots,\hat{y_{i-1},z}) = f(s_i,\hat{y_{i-1}})
\end{equation}

Where $y_{i-1}$ is the character embedding correspond to the last character generated, and $s_i$ is a state at step i representing the current processed information of both the molecule latent representation $z$, and the formerly generated $y_i's$ up to i-2.





Our goal is to create a molecule that is different from the original molecule $M$.
Intuitively, we wish to explore the chemical space around the molecule $M$.
Therefore, during generation process we introduce a diversity component noising the multidimensional Gaussian parameters used for sampling the hidden vector $z$. 
More formally, to introduce $diversity$ to our generation process, we instantiate our encoder output parameters with a diversity layer. Intuitively, the diversity layer outputs a noisy sample from a distribution centered as the encoder suggested, but with larger variance. This allows us to explore the molecule space around an origin molecule, with tune-able amount of diversity, corresponds to variability in chemical space. The diversity layer samples noisy instance according to the encoded Gaussian parameters and a diversity parameter $D$.

The output of the diversity layer is a sample from a conditional diverse distribution, described as follows: \\
Given the encoder outputs: vector of means $\hat{\mu_i}$ and standard deviations $\hat{\sigma_i}$ , and random noise sample n - from Gaussian distribution with diversity parameter D - [$n \sim \mathcal{N}(O,D)]$.

\vspace{-0.2cm}
\begin{equation}
\text{Diverse z} = (n \times \hat{\sigma_i}) + \hat{\mu_i} \sim \mathcal{N}(\hat{\mu_i}, \hat{\sigma_i}^2 \times D)
\label{equation:diversity}
\end{equation}

We obtain instance from the diverse distribution as our final noisy encoded representation ($z$) for the compound $M$, used as the base to for the decoder diversity-driven molecule generation.

We note that during training our diversity parameter D is set to 1. Thus $z$ instance is sampled from the non-diverse distribution suggested by the computed parameters.
Tuning this parameter at generation time allow us to explore the space around the prototype.






\vspace{-0.2cm}
\subsection{\NAME Architecture} \label{Architecture}

We leverage  recent advances in generative models and deep learning for natural language process (NLP) to form the prototype hypothesis generation process as an end-to-end deep neural network solution.
Figure \ref{fig:architecture} presents \NAME (Conditional Diversity Network) architecture.
\NAME starts by encoding the molecule (in SMILES notation) using the encoder function. First, encoding each character in the SMILES representation into its $d$ dimensional embedding, then applying convolutions over various substring (filter) sizes (e.g. correspond to chemical substructures). A similar encoder architecture was suggested for NLP tasks, such as sentence classification \cite{kim2014convolutional}.
The extracted features are then concatenated and fully connected layers are applied.
The outputs of the encoder are considered as a vector of means and a vector of standard deviations, representing the distributions of features for the prototype.
In VAE, those vectors are then fed into a decoder. The goal is to optimize reconstruction of the original input and constraint the representation to a known prior.
During generation, the vectors of features are sampled from the prior distribution and their output is passed to the decoder that generates a new representation.

We extend the VAE generation process by adding a \emph{diversity layer}.
During generation, instead of sampling from the prior means and standard deviations, we first feed a prototype.
We sample from the prototype feature distribution with parametrized diversity (Section \ref{dataGeneration}) to form the prototype latent representation -- served as input for the decoder.

As described in section \ref{dataGeneration}, our decoder is a sequential generator. By generating sequentially, we form another parameter of variability in the generated data, by introducing minor variations into the molecule generated during generation. 
This is the main component of many other works on molecule generation \cite{segler2017generating,insilicoRNN1,insilicoRNN2} to introduce diversity into the generation process.
We later show that our diversity layer can introduce diversity beyond this component.

We represent our decoder as a recurrent neural network (LSTM). The decoder receives the encoder output as its input. The encoded representation forms the first state of the decoder. The decoder then generates the compound sequentially (character by character) by operating on the distribution over characters at each time step, based on its updated state and the input character from former step.

During training, we feed the decoder with the correct next symbol, even if it was predicted wrongly \cite{teacherForce}. During generation, we experiment with two options for generating the next symbol: one by selecting the best scored character from distribution over symbols (argmax), and the second is by sampling from the same distribution. By introducing sampling into the generation, we are able to increase the amount of variability we generate during generation.
The model is trained to reconstruct the input prototype from a low dimensional continuous representation, by minimizing a discrete reconstruction loss. 
Formally, to minimize the reconstruction error on a discrete molecule representation, we use the cross-entropy loss, defined as:
\begin{equation}
H(y,\hat{y}) = \sum{y_i\log{\hat{y_i}}}
\end{equation}

We note that we minimize the variational lower bound \cite{kingmaVAE}, which is essentially optimizing the reconstruction error while constraining the latent distribution with a prior. 
To reconstruct syntactically valid SMILES, the generative model would have to learn the SMILES, which includes keeping track of atoms, rings and brackets to eventually close them. In this case, the lower dimension representation that can be reconstructed into a valid prototype, is a highly informative representation. In other words, by minimizing the reconstruction error, we want learn a prototype continuous representation that captures the coordinates along the main factors of variation in the chemical space. This representation is the base for further diversifying the molecule generation process.

\section{Experimental Settings}

In this section we provide details on the datasets, hyperparameter setting, and the training in general. Then, we mention the methods compared and used in our experiments.

\subsection{Model Details}

\NAME was trained using a Tensorflow API \cite{tensorflow}. We use the Adam algorithm \cite{kingma2014adam} to optimize all the parameters of the network jointly, regarding weights initialization - the atoms embedding were initialized using a random uniform distribution ranging from -0.1 to 0.1, convolution weights used truncated normal with std 0.1, all other weights used the Xavier initialization \cite{xavierInitialization}, biases were initialized with constant. To reduce overfitting, we included an early stopping criteria based on the validation set reconstruction error. We use exponential decay factor on learning rate, and the teacher forcing method \cite{teacherForce} during training. In total, table \ref{hyperParametersTable} presents \NAME hyper parameter configuration. 

The code for our system is available over github\footnote{\url{https://github.com/shaharharel/CDN_Molecule}} for further research in the community

\begin{table}[ht]
\centering
   \begin{tabular}{|c|c|}
    \hline
        Parameter & Value \\ \hline
        max molecule length & 50  \\ \hline
        char embedding size & 128  \\ \hline
        filter sizes & 3, 4, 5, 6 \\ \hline
        number of filters & 128 \\ \hline
        latent z dimension & 300  \\ \hline
        batch size & 64  \\ \hline
        initial learning rate  & 0.001  \\ \hline
        LSTM cell units & 150 \\ \hline
  \end{tabular}
\caption{\NAME hyperparameter configuration}
\label{hyperParametersTable}
\end{table}
\vspace{-0.7cm}
\subsection{Datasets} \label{datasets}

\subsubsection{Drug-like molecules database}
In our work, we provide experiments showing that CDN is capable of generating drug-like molecules. We train our model on a large drug-like molecules database and present several metrics on the generated molecules. 
The ZINC database \cite{zinc} contains commercially available compounds for structure based virtual screening. In addition, the database have subsets of ZINC filtered by physical properties. One such filtering is based on Lipinski's rule of five \cite{lipinski} -- a heuristic method to evaluate if a molecule can be a drug.
The subset contains over 10 million unique drug-like compounds.
\NAME was trained on a subset with approximately 200k drug-like compounds extracted at random from the ZINC drug-like database.
The subset was further divided to train/validation/test sets, with 5k compounds for validation and test sets, and the rest for training set.
The subsets are used for training the model (train), evaluating hyperparameters and stopping criteria (validation), and for method evaluation and experiments (test).

\subsubsection{Drug database}
For our drug-generation experiment (Section \ref{subsection:drugGeneration}) we show that some of the molecules generated by CDN are drugs which were discovered years later.
The DrugBank database \cite{drugBank} is a bioinformatics and cheminformatics resource that combines detailed drug data with comprehensive drug target information.
For retrospective experiments, we extracted a test set of 869 FDA approved drugs from the DrugBank database.
Note, our system is not trained on drugs, but rather presented with drug prototypes only during generation.

\subsection{Compared Methods} \label{baselines}
As discussed in Section \ref{relatedWork}, not much work has been done in the area of deep drug generation and specifically not on the diversity aspect of the generation. To the best of our knowledge, the works that do consider the task, do not aim at prototyping specific compound, but training for unconditional molecule generation and later apply post-processing to achieve general molecular characteristics. We compare our methods to the state of the art models for molecule generation on the reconstruction criteria, and further show that our model is able to build on top of those models to apply diversity. Specifically, we compare all following methods

\begin{enumerate}
\item $Seq2Seq$  \cite{seq2seq} -  An autoencoder architecture applied on sequence data for prediction of sequences. Both encoder and decoder are recurrent neural networks (RNN). Although the model is in general deterministic, it is able to bring stochasticity (and thus novelty) into the molecules generation process by setting the RNN decoder to sample from the distribution over characters in every time step instead of predicting the topmost next character. 
We therefore consider two baselines -- one using the Argmax method and the other utilizing the Sampling method to reach diversity.
\item $Conv2Seq$ - To conform better with \NAME parameter setting that utilize on CNNs, we implement a second auto encoder same as the previous method but with convolution encoder.
\item VAE \cite{kingmaVAE}-- a vanilla implementation of VAE. This model generates new molecules from unit Gaussian random samples, regardless of prototypes. 
\item \NAME-VAE - Our diversity model on top of variational auto encoder. $D$ is the diversity parameter of Equation \ref{equation:diversity}
The higher the $D$, the higher the diversity induced.
We note, that for $D=1$, the model extends VAE for a conditional setting but without diversity.
\end{enumerate}

\begin{table*}
    \centering
    \normalsize
    \begin{tabular}{|c|c|c|c|c|c|c|}
    	\hline
        Model & Acc & Valid & Novel & Acc @ 1k & Valid  @ 1k& Novel @ 1k\\ \hline
        Seq2Seq - Argmax & 0.94 & 0.93 & 0.13 & - & - & -\\ \hline
        Seq2Seq - Sampling & 0.91 & 0.88 & 0.19 & 0.92 & 0.89 & 32.5\\ \hline
        Conv2Seq - Argmax & 0.92 & 0.85 & 0.14 & - & - & -\\ \hline
        Conv2Seq - Sampling & 0.89 & 0.77 & 0.18 & 0.88 & 0.76 & 35.2\\ \hline
        VAE & -\footnotemark & 0.58 & - & - & - & - \\ \hline

        \NAME - D=1 & 0.91 & 0.89 & 0.19 & 0.9 & 0.89 & 8 \\ \hline
        \NAME - D=2 & 0.82 & 0.81 & \textbf{0.26} & 0.81 & 0.8 & \textbf{66.6}\\ \hline
        \NAME - D=3 & 0.64 & 0.63 & \textbf{0.37} & 0.65 & 0.65 & \textbf{227}\\ \hline

   \end{tabular}
   \caption{Evaluation of \NAME and baselines for diversity and validity of generated molecules}
   \label{table:druglikeEval}
      \vspace{-0.4cm}
\end{table*}

\begin{table*}
\centering
   \begin{tabular}{|c|c|c|c|}
    \hline
     Input Drug& Input SMILES& Generated Drug& Generated SMILES\\ \hline
     Aminosalicylic & \small $Nc1ccc(C(=O)O)c(O)c1 $ & Mesalazine & \small $Nc1ccc(O)c(C(=O)O)c1$ \\ \hline
     Pyrazinamide & \small $NC(=O)c1cnccn1$ & Isoniazid & \small $NNC(=O)c1ccncc1$ \\ \hline 
     Protriptyline & \small $CNCCCC1c2ccccc2C=Cc2ccccc21$ & Desipramine & \small $CNCCCN1c2ccccc2CCc2ccccc21$ \\ \hline
     Phenelzine & \small $NNCCc1ccccc1$ & Isoniazid & \small $NNC(=O)c1ccncc1$ \\ \hline
     Isoproterenol & \small $CC(C)NCC(O)c1ccc(O)c(O)c1$ & Orciprenaline & \small $CC(C)NCC(O)c1cc(O)cc(O)c1$ \\ \hline 
     Pheniramine & \small $CN(C)CCC(c1ccccc1)c1ccccn1$ & Tripelennamine & \small $CN(C)CCN(Cc1ccccc1)c1ccccn1$ \\ \hline 
  \end{tabular}
\caption{Sample of automatically generated drugs and the drug served as prototype to the generation process}
\vspace{-0.4cm}
\label{generatedDrugsTable}
\end{table*}

\section{Experiments}
\label{s:Experiments}
In this section, we first conduct several experiments to determine \NAME performance in the task of reconstructing the molecular structure. We present evaluation of the  trade off between the molecules reconstruction accuracy and novelty as a function of \NAME diversity component.
Additionally, we conduct several drug related experiments to show \NAME capabilities in the real world for generating new drugs.

\subsection{Novel Molecules Generation} 
\label{subsection:novelMoleculeGeneration}
Our main goal is to create novel molecules that carry similarities to the prototype. Thus, the metric of reconstruction is an important metric. 
We examine the methods on the task of prototype reconstruction on a test set of 5k ZINC drug-like compounds.
To explicitly address the reconstruction accuracy and validity vs. the generated molecules diversity, we measure the following metrics:
\begin{enumerate}
\item \emph{Reconstruction Accuracy (Acc)} - Character-level accuracy with the input prototype served also as target.
\item \emph{Valid Molecule Percentage (Valid)} - Percentage of valid molecules. 
There are several numeric validations performed on molecules representation to validate its correctness. We used Rdkit \cite{rdkit} library to measure validity of the generated compounds.
\item \emph{Novel Molecule Percentage (Novel)} - A novel molecule is both a valid molecule, and different from the prototype.
\end{enumerate}
\vspace{-0.1cm}
To be able to measure the molecule generation capabilities over various Gaussian samples for the same prototype compound (we want to be able to generate \emph{several} compounds related to the origin compound), we also measure all the above metrics with $@k$ notation. In our context, $@k$ represents that for each prototype compound, we run \NAME generation process with $k$ instances of random noises parametrized with diversity $D$.
We note that to measure $Novel@k$, we count how many unique molecules generated -- a novel molecule is counted only once, even if it was generated with various Gaussian samples for the prototype. The metric of $Novel@k$ is not normalized, thus, the semantics of this metric should be, intuitively, interpreted as how many unique molecules were generated for a prototype and 1000 Gaussian samples. 

Table \ref{table:druglikeEval} presents the results of \NAME and the baselines on the metrics  above. 
Analyzing $Acc$, $Valid$ and $Novel$ metrics, we observe that 
with diversity level of 1 (non-diverse sampling), \NAME generates similar diversity to the baselines. Increasing $D$, significantly increases the diversity in the generated molecules, while reducing the level of accuracy and the valid molecules rate.
This result stems from the intuition that as the representation becomes noisier, it is harder for the model to reconstruct the original prototype.
Adressing the $@k$ metrics, we observe \NAME is able to maintain the accuracy and validity levels with many random samples used for generation, while generating various unique molecules for the same prototype input, with the number of unique molecule significantly increasing with the diversity parameter $D$.
\footnotetext{As no prototype is given, there is no reconstruction to measure}

\subsection{Drug Generation}
\label{subsection:drugGeneration}
The main aim of this work is to generate novel molecules with desired properties (characterized by the prototype molecule), by searching the chemical space around the prototype.
To check the immediate benefits (i.e., without further screening the generated compound) of our approach to a real world task, we conduct a retrospective experiment in the drug domain. We apply our method on a test set of FDA approved drugs as prototypes. We note that none of the drugs was observed in the training data, which was composed of only drug-like molecules.

Evaluation on this task is harder since our goal is to generate drugs, and we cannot a-priori know if the generated molecule has the desired characteristics of a drug without further experimenting with the compound. We therefore consider as gold standard a test set of 869 approved known drugs. Although this test set is very small in compare with the enormous molecule space, some approved drugs are chemically similar and share similar therapeutic characteristics, thus we hypothesize that by applying \NAME on FDA approved drugs as prototypes, we might be able to generate other known compounds / drugs with similar characteristics.

Interestingly, targeting some existing drugs as prototypes, our model was able to generate molecules that also appear in the FDA approved drugs list and are closely related to the prototype, both in the chemical aspects and by their medical use (i.e. targeting the same biological mechanism of action). Table \ref{generatedDrugsTable} presents a sample of the drugs generated.

In total, we run the baselines and \NAME variants over all 869 approved drugs dataset as prototypes, with 1000 Gaussian samples in each run. Table \ref{table:drugGeneration} presents the number of FDA approved generated drugs with each method. We also present the percentage those drugs constitute from the valid molecules generated. 
We draw the reader attention to the negligible chance of generating a drug using exhaustive search without constraints (e.g., using HTS). 
We observe that the VAE could not produce any known drug. We hypothesize that this stems from the fact that VAE randomly generates a molecule and not based on a prototype. 
\NAME with no diversity and the other baselines generated 9--12 drugs.
This result emphasize how the variability that the decoders present during sampling contributes to the generation of known drugs. 
More interestingly, we observe that for higher $D$ values of our diversity layer (CDN-VAE D=2 and CDN-VAE D=3), the amount of known drugs increases significantly.
One should remember that the model doesn't have any ``drug'' understanding -- the model was only trained given drug-like molecules, and \emph{all known drugs were eliminated from the training}.
The key here is the chemical similarity drugs share. Thus, by targeting a drug molecule as prototype to the generation process, our model is able to chemically diversify the prototype drug in a way that generated another known drugs.
We are encouraged by the results that \NAME was able to generate a significant number of already known drugs.
We are currently testing with pharmaceuticals companies the additional generated molecules. 

\begin{table}
    \centering
    \begin{tabular}{|c|c|c|}
    	\hline
        Model& \#Drugs & \vtop{\hbox{\strut \% from Generated }\hbox{\strut Valid Molecules}} \\ \hline
        VAE & 0 & 0\%\\ \hline
        Seq2Seq & 12 & 0.002\% \\ \hline 
        Conv2Seq & 9 & 0.0018\% \\ \hline 
        CDN-VAE D=1& 12& 0.0023\% \\ \hline 
        CDN-VAE D=2& 22& 0.005\% \\ \hline
        CDN-VAE D=3 & 35 & \bf{0.01\%}\\ \hline 
   \end{tabular}
   \caption{Automatically generated FDA approved drugs. We present the percentage of the FDA-approved drug from the total valid molecules generated by each method}
   \label{table:drugGeneration}
   \vspace{-0.8cm}
\end{table}

\subsection{Qualitative Examples}
We present a few qualitative examples of the drugs generated.
We would like to explore whether the application of the system on drugs developed up until a certain year might find drugs that will be discovered years later.
During training we eliminate all known drugs from the ZINC database and we present as prototypes a single drug.
Figure \ref{fig:timeline} presents a timeline with example pairs of origin (top row) and generated (bottom row) molecules, with the year of the drug first use. 
By using \NAME we could have generated the bottom molecules directly when we knew the origin molecules, possibly sparing years of research.
The system was able to identify the main drug for Tuberculosis -- Isoniazid using an initial prototype of the disease that was never used due to its side effects (Pyrazinamide).
Additional intriguing example is the generation of Orciprenaline which is used to treat Asthma from a prototype drug that was mainly used for heart block, and very rarely for asthma.
These pairs are closely related in their therapeutic effect, but a few changes for the molecule were needed to reposition it for Asthma treatment.
Another interesting discovery was Mesalazine, used to treat inflammatory bowel disease based on an antibiotic primarily used to treat tuberculosis. discovered about 40 years before.



\subsection{Diversity Mechanisms}
A common method to employ diversity in encoder-decoder models is to employ a sampling decoder into the architecture \cite{segler2017generating,insilicoRNN1,insilicoRNN2}.
The diversity is introduced by sampling from distribution over characters in each time step of generation, rather than choosing the topmost (argmax) character at test time.
We analyze the contribution of the diversity layer $D$ for \NAME presented in this work along side a sampling decoder as well.
Table \ref{table:samplingDruglikeEval} presents \NAME performance on the previous metrics (Section \ref{subsection:novelMoleculeGeneration}), but with a sampling decoder. 
We compare \NAME with sampling but with no diversity component ($D=1$) to \NAME with sampling with higher values of $D$ and observe that the diversity parameter is able to introduce additional diversity beyond the sampling decoder component.

\begin{figure}
\centering
\begin{subfigure}[b]{0.38\textwidth}
   \includegraphics[width=1\linewidth]{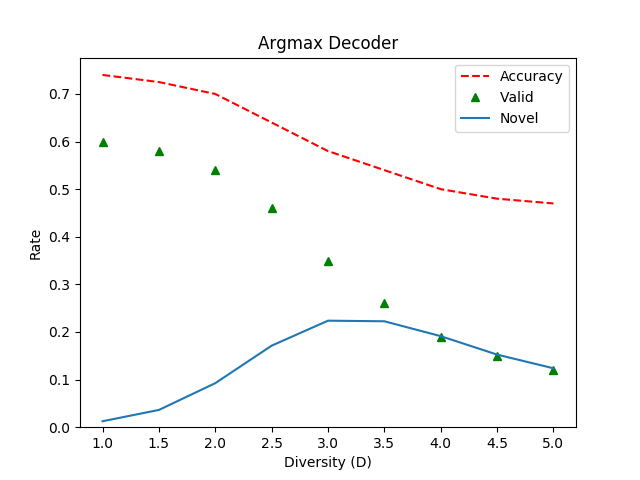}
\end{subfigure}
\begin{subfigure}[b]{0.38\textwidth}
   \includegraphics[width=1\linewidth]{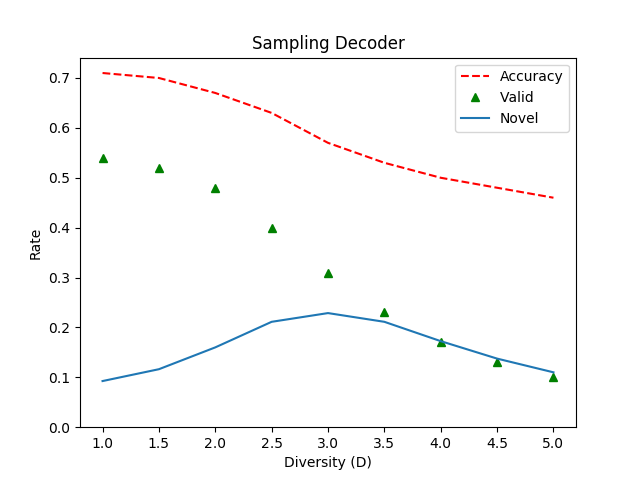}
\end{subfigure}
\caption{Diversity parameter effect on performance.}
\label{fig:diversity_param}
\end{figure}

\begin{table}
    \centering
    \begin{tabular}{|c|c|c|c|c|c|c|}
    	\hline
        Model & Acc & Valid & Novel & A@1k & V@1k& N@1k\\ \hline
       	\NAME D=1 & 0.88 & 0.78 & 0.19 & 0.88 & 0.79 & 39.5\\ \hline 
      	\NAME D=2 & 0.8 & 0.69 & \textbf{0.25} & 0.79 & 0.68 & \textbf{94}\\ \hline 
       \NAME D=3 & 0.57 & 0.39 & \textbf{0.27} & 0.56 & 0.38 & \textbf{179} \\ \hline

   \end{tabular}
   \caption{\NAME performance using a sampling decoder.}
   \label{table:samplingDruglikeEval}
   \vspace{-0.3cm}
\end{table}

%

\begin{figure*}
  \centering
  \begin{minipage}[b]{0.33\textwidth}
    \includegraphics[width=\textwidth,height=0.6\textwidth]{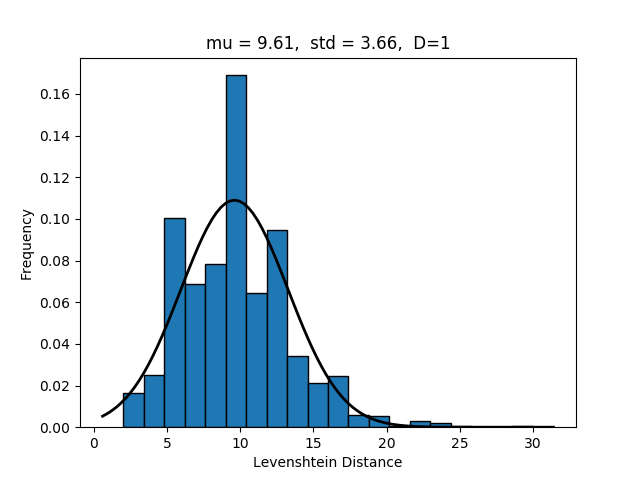}
  \end{minipage}
  \hfill
  \begin{minipage}[b]{0.33\textwidth}
    \includegraphics[width=\textwidth,height=0.6\textwidth]{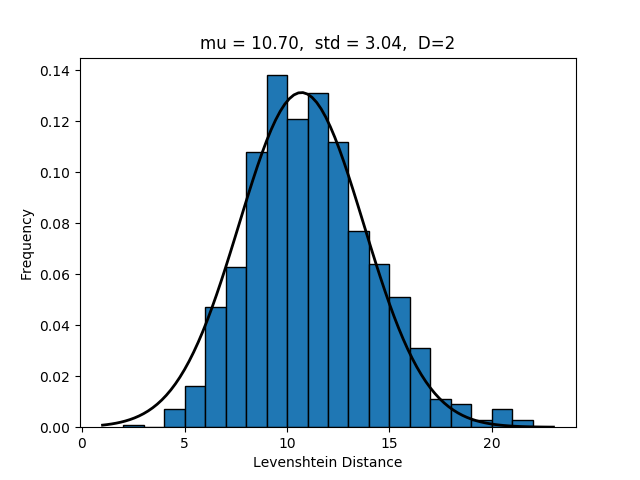}
  \end{minipage}
    \hfill
  \begin{minipage}[b]{0.33\textwidth}
   \includegraphics[width=\textwidth,height=0.6\textwidth]{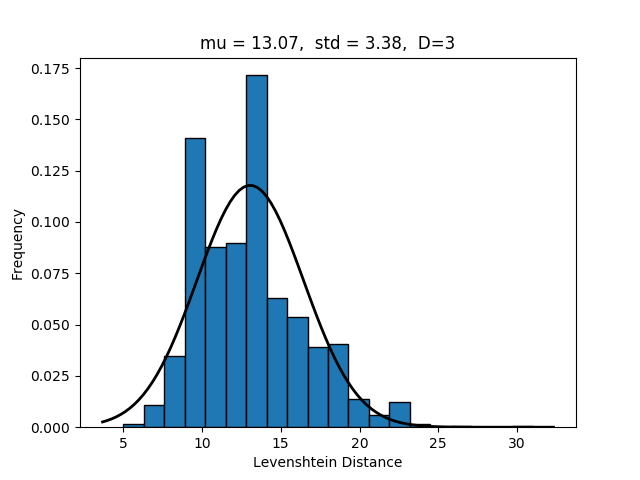}
  \end{minipage}
    \hfill
  \centering
  \begin{minipage}[b]{0.33\textwidth}
    \includegraphics[width=\textwidth,height=0.6\textwidth]{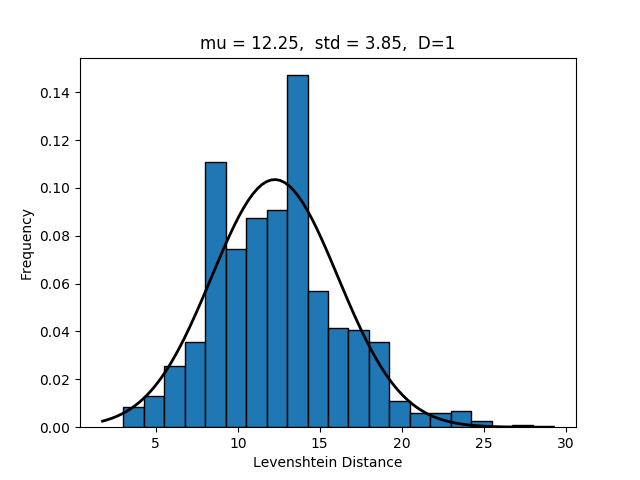}
  \end{minipage}
  \hfill
  \begin{minipage}[b]{0.33\textwidth}
    \includegraphics[width=\textwidth,height=0.6\textwidth]{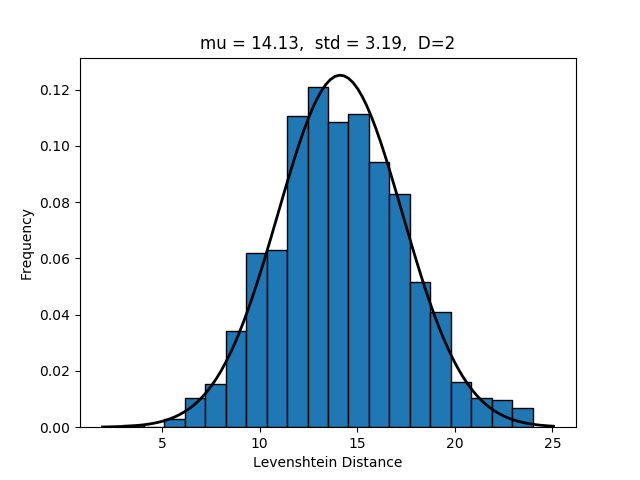}
  \end{minipage}
    \hfill
  \begin{minipage}[b]{0.33\textwidth}
    \includegraphics[width=\textwidth,height=0.6\textwidth]{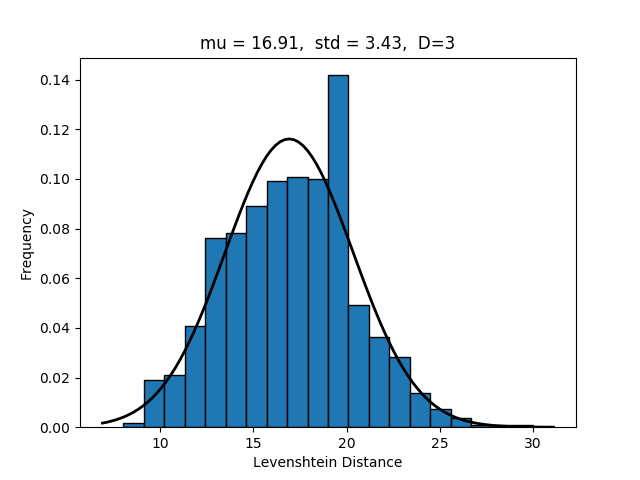}
  \end{minipage}
    \hfill
  \vspace{-0.5cm}
  \caption{Levenshtein distance histograms for analyzing the diversity generated by \NAME. Top - Origin molecule vs Generated molecule distances. Bottom - within generated molecule population distances.}
   \label{fig:distanceHistograms}
\end{figure*}

To analyze the behavior of the diversity parameter $D$ on the accuracy/validity and novelty trade-off in the drug domain, we generated samples for the FDA approved test set (Section \ref{datasets}), with various configurations of the diversity parameter D.
Figure \ref{fig:diversity_param} presents the results for the two types of decoder functions.
As we hypothesized, with both decoders, increasing the value of the diversity parameter $D$, significantly increases the amount of novel molecules generated. As we expected, the novelty is not free, we observe lower accuracy and lower valid rates for increased diversity.
Comparing argmax with sampling decoders, we observe that in general, sampling has lower accuracy and valid rate, but for low diversity value the sampling method generates significantly more novelty than the argmax. This behavior reduces for higher values of the diversity parameter, were both methods generates similar rates of novelty.
We also observe the novelty rate reduces at some point of increased diversity value. This is quite expected because for large values of diversity, the latent molecule representation sampled with larger noise, thus at some point the generator is not able to recover much valid molecules in general, and novel ones in particular.

\vspace{-0.2cm}
\subsection{Molecular Variations}
We would like to analyze not only whether a molecule is different from the prototype molecule but also quantify the diversity of the molecules with respect to the prototype molecule.
Additionally, we would like to validate that the generated molecules originating from a prototype are also diverse with respect to one another.
We compare the Levenshtein distances of the generated SMILES within the generated population and with respect to the prototype, used as input for a specific generation instance.
We apply \NAME on the drug-like test set as prototypes. We note that we count only valid molecules generated in all evaluations.
Figure \ref{fig:distanceHistograms} presents histograms of the Levenshtein distances for the generated molecules, with approximated Gaussian parameters and curve on top of the histograms. The top row represents the input prototype compared to the generated molecules Levenshtein distance distribution for different configurations of the diversity parameter $D$ (increasing $D$ from left to right). 
The bottom row represents the inner-generated population Levenshtein distance distribution for various values of $D$. 
On both type of distance evaluations (rows), we observe significantly larger Levenshtein distances for larger values of $D$, thus indicating positive effect of the diversity parameter $D$ on both the distance from the prototype molecule, and the average inner distance between molecules that were generated with different random samples to the same prototype. 
Additionally, we observe \NAME diversity in generation is not limited to generating diversity with respect to the origin molecule, but also generates diversity within the generated population for a specific prototype, with higher amount of diversity tuned with the diversity parameter D.

\begin{table}
    \centering
    \begin{tabular}{|c|c|c|c|}
    	\hline
        Class&Cosine&L2&L1\\ \hline
        Thiazide Diuretics & 0.872 & .95 & .908 \\ \hline
        Benzodiazepines & 0.923 & .883 & .859 \\ \hline 
        $\beta$-Blocker & 0.866 & .849 & .822 \\ \hline 
        NSAIDs & 0.955 & .853 & .833 \\ \hline \hline
        Across Drugs & 1.0 & 1.0 & 1.0 \\ \hline 
   \end{tabular}
   \caption{In class and across drugs normalized distances computed on various drug classes.}
   \label{drugClassesTable}
   \vspace{-0.5cm}
\end{table}



\subsection{Molecule Representation in Latent Space}
Encoder-decoder settings produce intermediate representations of their input.
In this section, we analyze the quality of those representations.
%
During \NAME generation process, we first encode molecule into a low dimensional vector space with the encoder function. We refer to the output vector as the molecule embedding.
To evaluate the embeddings, we leverage them for the task of drug classification. Intuitively, if the embeddings captures enough information for drug classification, we might rely on this representation for molecule generation.
We note that for the task of encoding the molecule feature representation, we set the diversity parameter $D=1$, but one should remember that the representation is still instantiated from unit Gaussian, and thus is not deterministic.

A drug class is a set of medications that have similar chemical structures, or the same mechanism of action (i.e., bind to the same biological target).
In Table \ref{drugClassesTable} we report embedding vector normalized distances in-class and across various drug classes.
Thiazide and Benzodiazepines are chemical classes while $\beta-Blocker$ and NSAIDs \footnote{NSAIDs - Nonsteroidal anti-inflammatory agents} are classes representing mechanism of action. We observe all in-classes distances are significantly lower than across class. We conclude that although our molecule representation is noisy by the stochastic nature of \NAME, similarities in the embedding space are able to reflect significant similarities among various drug class.

\section{Conclusions}
\vspace{-0.1cm}
Drug discovery is the process of identifying potential molecules that can be targeted for drugs.
Common methods include systematic generation and testing of molecules via HTS. However, the molecular space is very large.
Additional approaches require chemist to identify potential drugs based on their knowledge. Usually, they would start from a known compound in nature or known drug and identify potential changes.
Approaches in machine learning today mainly focused on non-controlled molecule generation using generative mechanisms, such as VAE.
The approaches were limited in their ability to generate both valid and novel molecules.
In this work, we presented a prototype-based approach for generating drug-like molecules. 
We adopt the chemist approach of ``borrowing'' from nature or focusing on known drugs.
We hypothesize that biasing the molecule generation towards known drugs might yield valid molecules.
We train our model on drug-like molecules, and during generation extend VAE to, intuitively, search closer to the prototype (which can be a drug).
We add additional component to diversify the molecules generated.
We present results that show that many of the molecules generated are both valid and novel.
When conditioning on drugs, we observe our system was able to generate known drugs that it never encountered before.
The system is currently being deployed for use in collaboration with pharmaceutical companies to further analyze the additional generated molecules.
\vspace{-0.2cm}

\bibliographystyle{ACM-Reference-Format}
\bibliography{acmart} 

\end{document}